\documentclass{llncs}
\usepackage{makeidx}  
\usepackage{graphicx}
\begin{document}

\mainmatter              
\title{Magical Number Seven Plus or Minus Two: 
Syntactic Structure Recognition 
in Japanese and English Sentences}
\titlerunning{Magical Number Seven Plus or Minus Two}  
%
\author{Masaki Murata \and Kiyotaka Uchimoto \and Qing Ma \and Hitoshi Isahara}

\authorrunning{Masaki Murata et al.}   
%
\tocauthor{Masaki Murata(Communications Research Laboratory, MPT),
Kiyotaka Uchimoto(Communications Research Laboratory, MPT),
Qing Ma(Communications Research Laboratory, MPT),
Hitoshi Isahara(Communications Research Laboratory, MPT)}

\institute{Communications Research Laboratory, MPT,\\ 
2-2-2 Hikaridai, Seika-cho, Soraku-gun, Kyoto, 619-0289, Japan,\\
\email{\{murata,uchimoto,qma,isahara\}@crl.go.jp},\\ WWW home page:
\texttt{http://www-karc.crl.go.jp/ips/murata}}

\maketitle              

\def\None#1{}

\begin{abstract}
George A. Miller said that 
human beings have only 
seven chunks in short-term memory, plus or minus two. 
We counted the number of 
bunsetsus (phrases) whose modifiees are undetermined 
in each step of an analysis of 
the dependency structure of Japanese sentences, 
and which therefore must be stored in short-term memory. 
The number was roughly less than nine, 
the upper bound of seven plus or minus two. 
We also obtained similar results 
with English sentences under the assumption 
that human beings recognize 
a series of words, such as a noun phrase (NP),  
as a unit. 
This indicates that 
if we assume that the human cognitive units 
in Japanese and English are 
bunsetsu and NP respectively, 
analysis will support Miller's $7 \pm 2$ theory. 
\end{abstract}
%

\section{Introduction}
George A. Miller suggested in 1956 that 
human beings have only 
seven chunks\footnote{A chunk is a cognitive unit of information.}
in short-term memory, plus or minus two \cite{miller56}. 
We counted the number of 
{\it bunsetsus} (phrases) 
whose modifiees are undetermined 
in each step of an analysis of 
the dependency structure of Japanese sentences 
and which therefore must be stored in short-term memory, 
using the Kyoto University corpus \cite{kurohashi:nlp97_e}. 
(The Kyoto University corpus is a syntactic-tagged corpus 
collected from editions of the Mainichi newspaper.) 
The number was roughly less than nine, that is, 
the upper bound of Miller's $7 \pm 2$ rule. 
This result supposes that 
bunsetsus whose modifiees are not determined are 
stored in short-term memory. 
For the Kyoto University corpus, 
the number of stored items was less than nine. 
This result supports Miller's theory. 
We made a similar investigation of English sentences 
using a method described by Yngve \cite{yngve60}. 
We assumed that human beings 
recognize a series of words, such as a noun phrase (NP), 
as a unit and found that 
the required capacity of short-term memory is 
roughly less than nine. 

\section{Short-term memory and the $7 \pm 2$ theory}

Miller said that 
human beings have only 
seven chunks in short-term memory, plus or minus two, 
because the results of various experiments 
on words, tones, tastes, sight organs 
indicated approximately seven. 
The ``plus or minus two'' indicates an individual-based 
variation\footnote{
Note that the following descriptions are not 
directly related to Miller's $7 \pm 2$ theory, 
but to short-term memory. 
Lewis's work, ``Magical number two or three,'' 
discussed linguistic features 
related to short-term memory \cite{Lewis96}. 
The work discussed the number of 
center-embedded sentences and 
theorized that in English 
only one main clause sentence and one center-embedded sentence, 
for a total of two sentences, are allowed. 
In Japanese, 
one main clause sentence and two center-embedded sentences, 
for a total of three sentences, are allowed. 
These limitations are caused by the constraints of short-term memory, 
and have been discussed in English 
in principle four, ``Two sentences'', of 
Kimball's Seven Principles \cite{Kimball73}. 
This research suggests that 
the reason for the limited number of 
center-embedded sentences is 
the limited capacity of human short-term memory.}. 

Although the research on the $7 \pm 2$ theory belongs 
to the field of psychology, 
it can be applied to the field of engineering. 
In sentence generation, for example, 
a sentence that exceeds the seven plus or minus two capacity of 
short-term memory is difficult to understand, 
so sentences are generated that do not exceed 
this upper limitation \cite{yngve60}. 
In human-interface systems, 
only about seven plus or minus two objects are 
displayed at one time 
because if more pieces of information are given, 
humans have trouble recognizing the images. 
Research on the $7 \pm 2$ theory is useful not only for 
the scientific investigation of human beings, 
but also for the engineering of things used in daily life. 

\section{Investigation of Japanese sentences}

\begin{figure*}[t]
      \begin{center}
      \includegraphics[height=6cm,width=12cm]{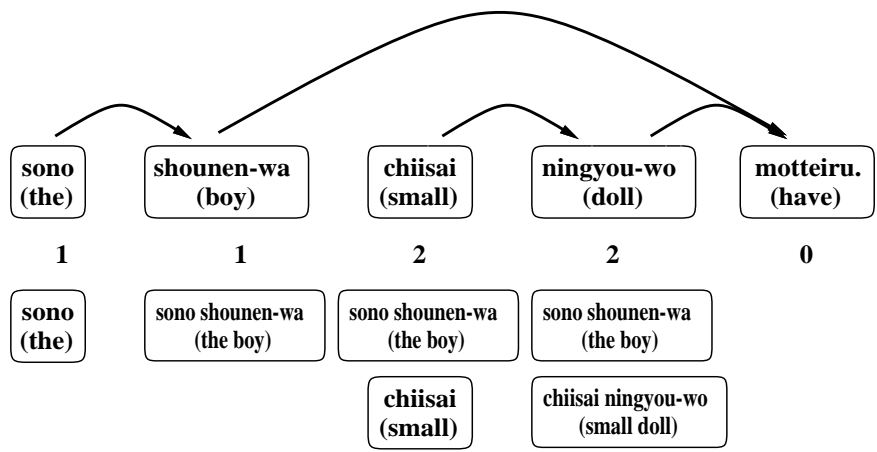} 
      \end{center}
    \caption{How to estimate the number of bunsetsus 
      whose modifiees are not determined}
    \label{fig:jap}
\end{figure*}

In this work, 
we consider the process of sentence understanding 
as the analysis of the syntactic structure of a sentence, 
and we assume that those items which must 
be stored in short-term memory when understanding a sentence 
are bunsetsus whose modifiees are not determined. 
(``Bunsetsu'' is a Japanese technical grammatical term. 
A bunsetsu is like a phrase in English, 
but it is a slightly smaller component. 
{\it Eki-de} ``at the station'' is a bunsetsu, and 
{\it sono}, which means ``the'' or ``its,'' is also a bunsetsu. 
A bunsetsu is roughly a unit referring to an entity. 
So a bunsetsu is thought to be an appropriate unit of recognition.) 
Figure \ref{fig:jap} is an example of calculating 
the number of bunsetsus whose modifiees are not determined 
in each step when analyzing 
the syntactic structure of the following sentence 
from left to right. 

\begin{tabular}[h]{@{ }c@{ }c@{ }c@{ }c@{ }c}
& & &\\
{\it sono} & {\it shounen-wa} & {\it chiisai} & {\it ningyou-wo} & {\it motteiru.} \\
(the) & (boy) & (small) & (doll) & (have) \\
\multicolumn{5}{l}{The boy has a small doll.}\\
& & &\\
\end{tabular}

Arrows in the figure indicate the dependency structure. 
The number indicates the number of bunsetsus 
whose modifiees are not determined, and 
the lower part indicates the elements 
which must be stored in short-term memory. 
At the beginning, 
when {\it sono} (the) is input, 
its modifiee has not been determined yet, 
so it must be remembered. 
It is then stored in short-term memory 
as a bunsetsu whose modifiee is not determined. 
When {\it shounen} (boy) is input, 
{\it sono} (the) is found to modify 
{\it shounen} (boy). 
So {\it sono} (the) will not be used 
in the syntactic analysis after that, and 
it does not need to be remembered independently. 
{\it Sono} (the) is recognized  
to be attached to {\it shounen} (boy) 
in the form of {\it sono shounen} (the boy). 
As a result, only one element, {\it sono shounen} (the boy), 
whose modifiee is not determined, 
is stored in short-term memory. 
Next, {\it chiisai} (small) is input. 
This time, 
the dependency structure is not changed, 
and {\it sono shounen} (the boy) and 
{\it chiisai} (small) are stored in short-term memory. 
Next, when {\it ningyou} (doll) is input, 
{\it chiisai} (small) is recognized 
to modify {\it ningyou} (doll). 
{\it Chiisai} (small) will not be used 
in later analysis, because it is recognized 
to be attached to {\it ningyou} (doll) 
in the form of {\it chiisai ningyou} (small doll).
Only the two elements 
{\it sono shounen} (the boy) and 
{\it chiisai ningyou} (small doll) are stored. 
Finally, {\it motteiru} (have) is input. 
Here, all the relationships of the dependency structure 
are determined and 
the number of bunsetsus with undetermined modifiees 
is 0. 
All the elements which were stored in short-term memory 
are cleared. 

\begin{table*}[t]
  \caption{Number of bunsetsus which undetermined modifiees}
  \label{tab:hindo_toukei}
  \begin{center}
\begin{tabular}[c]{|c|r|r|}\hline 
Bunsetsus with un-  & \multicolumn{2}{c|}{Frequency}\\\cline{2-3}
determined modifiees    &   \multicolumn{1}{c|}{Bunsetsu} & \multicolumn{1}{c|}{Sentence}\\\hline
  0 &      19954 &         90\\
  1 &      52751 &       1352\\
  2 &      59494 &       5022\\
  3 &      38465 &       6823\\
  4 &      15802 &       4468\\
  5 &       4488 &       1593\\
  6 &       1143 &        480\\
  7 &        195 &        102\\
  8 &         47 &         17\\
  9 &         10 &          5\\
 10 &          3 &          2\\\hline
\end{tabular}
\end{center}
\end{table*}

We assume that 
all human beings understand sentences the above way. 
The results are shown in Table \ref{tab:hindo_toukei}. 
The number in the ``bunsetsu'' column 
is the number of bunsetsus 
having the given number of undetermined modifiees 
among all the bunsetsus of the Kyoto University corpus, 
(19,954 sentences and 192,352 bunsetsus). 
The number in the ``sentence'' column 
is the number of sentences having 
the given number of 
undetermined modifiees. 
In this table, 
only three bunsetsus exceeded the upper bound of 
Miller's $7 \pm 2$ rule. 
The result supports Miller's theory. 

\None{
We counted the number of 
bunsetsus with undetermined modifiees 
in the Kyoto University corpus 
to create the definition of dependency structures. 
But this has some problems. 
For example, we included 
bunsetsus such as conjunctions, 
which may not need to be stored. 
If we appropriately handle such problems, 
the number of bunsetsus 
with undetermined modifiees should decrease. }

\section{Investigation of English sentences}

In the investigation of the Japanese corpus in the previous section 
we estimated the upper bound of the short-term memory 
required for sentence understanding. 
This section describes 
a similar investigation of an English corpus. 

\None{
At first, we described the method
counting the capacity of short term memory 
in English corpus, which 
was proposed by Yngve. 
Next, we showed Sampson's investigation 
in SUSSANNE corpus and 
reported our investigation in Penn Treebank. }

\begin{figure*}[t]
  \begin{center}
\fbox{
    \begin{minipage}{10cm}
      \begin{center}
      \includegraphics[height=8.25cm,width=7.5cm]{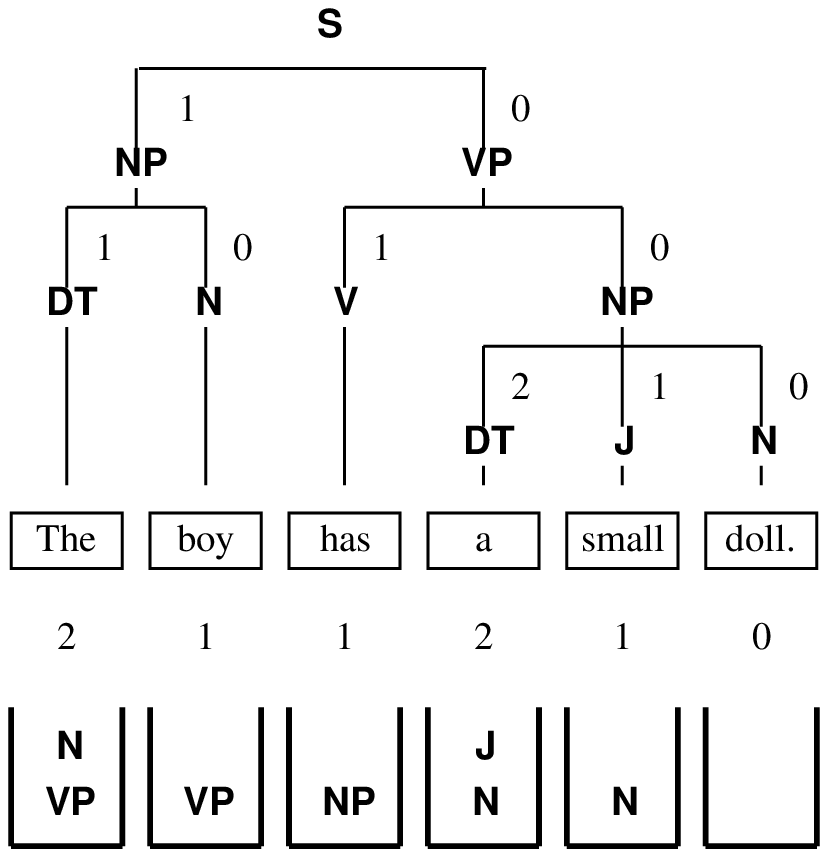} 
      \end{center}
      \vspace{-0.5cm}
    \caption{How to count the number of nonterminal symbols 
      stored in stacks}
    \label{fig:eng}
    \end{minipage}
}
  \end{center}
\end{figure*}

Yngve described a method for estimating the short-term memory capacity 
required in the syntactic analysis of 
an English sentence \cite{yngve60}. 
This method supposes that 
the nonterminal symbols, i.e., S and NP, 
which are stored in a stack when analyzing a sentence 
in a top-down fashion by using a push-down automaton, 
are those which need to be stored in short-term memory, and 
it counts the number of symbols stored in the stack. 
Figure \ref{fig:eng} shows 
how the number of nonterminal symbols stored in a stack 
is counted in the analysis of the sentence, 
``The boy has a small doll,'' in a push-down automaton. 
\begin{figure}[t]
  \begin{center}
\fbox{
    \begin{minipage}{7cm}
      \begin{center}
      \includegraphics[height=0.5cm,width=2cm]{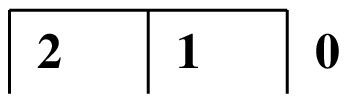}
      \end{center}
      \vspace{-0.5cm}
    \caption{How to give a number to each branch}
    \label{fig:eng_1}
    \end{minipage}
}
  \end{center}
\end{figure}%
\begin{figure*}[t]
  \begin{center}
\fbox{
    \begin{minipage}{10cm}
      \begin{center}
      \includegraphics[height=3cm,width=5cm]{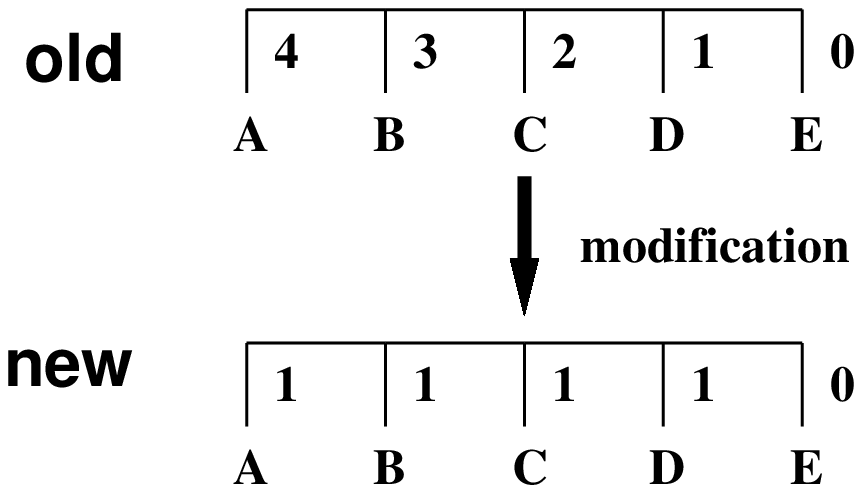}
      \end{center}
      \vspace{-0.5cm}
    \caption{Sampson's counting method}
    \label{fig:eng_3}
    \end{minipage}
}
  \end{center}
\end{figure*}%
Boxes in the lower part of Figure \ref{fig:eng} indicate 
the state of the stack as the sentence is parsed. 
For example, at the beginning of the sentence, 
``The'' is input first. 
When the sentence is analyzing in a top-down fashion, 
S is given first. 
Next, S is transformed into (NP VP). 
When VP is remembered, NP is transformed into (DT N). 
When N is remembered, 
DT is recognized to be ``The''\footnote{
Yngve's method has the following two problems. 
The first is, in Figure \ref{fig:eng}, 
we can select two possible patterns, (DT N) and (DT J N), 
in transforming NP, and we cannot select one of them 
when ``The'' is input. 
The other is that, 
by changing the grammar used in a corpus, 
the structure of a syntactic tree is changed and 
the result is changed. 
Despite these problems, 
we used Yngve's method because it is very easy to count with.}. 
As a result, 
the two non-terminal symbols, VP and N, 
need to be stored in a stack 
while ``The'' is processed. 
Similarly, 
the non-terminal symbols which 
need to be stored for each stack 
are shown in Figure \ref{fig:eng}. 
The numbers of symbols in the stacks 
for each word are 2, 1, 1, 2, 1, and 0. 
Yngve also proposed an easy method of counting 
the number of nonterminal symbols stored in a given stack. 
In this method a number is assigned 
to each branch of a tree as shown in Figure \ref{fig:eng_1}. 
The sum of the numbers in the path from 
S to a word is considered 
as the number of symbols stored in a stack at that word. 
For example, 
at the word ``The'', 
``1, 1'' is in the path of S, NP, DT, and ``The'', 
so the sum is 2, which matches 
the number of symbols stored in the stack. 

\begin{table*}[t]
  \caption{Number of nonterminals stored in a stack (SUSANNE corpus)}
  \label{tab:hindo_toukei_eng_sussane}
  \begin{center}
\begin{tabular}[t]{|r|r|}
\multicolumn{2}{c}{(a)Yngve's method}\\\hline
\multicolumn{1}{|l|}{Stack} & \multicolumn{1}{c|}{Frequency}\\
   &   \multicolumn{1}{c|}{(words)} \\\hline
0 & 7851 \\
1 & 30798 \\
2 & 34352 \\
3 & 26459 \\
4 & 16753 \\
5 & 9463 \\
6 & 4803 \\
7 & 2125\\
8 & 863\\
9 & 313\\
10 & 119\\
11 & 32\\
12 & 4\\
13 & 1\\\hline
\end{tabular}
\begin{tabular}[t]{|r|r|}
\multicolumn{2}{c}{(b)Sampson's method}\\\hline
\multicolumn{1}{|l|}{Stack}& \multicolumn{1}{c|}{Frequency}\\
    &   \multicolumn{1}{c|}{(words)} \\\hline
0 & 55866\\
1 & 64552\\
2 & 12164\\
3 & 1274\\
4 & 76\\
5 & 4\\\hline
\end{tabular}
\end{center}
\end{table*}

\begin{figure*}[p]
  \begin{center}
\fbox{
    \begin{minipage}{10cm}
      \begin{center}
      \includegraphics[height=3cm,width=5cm]{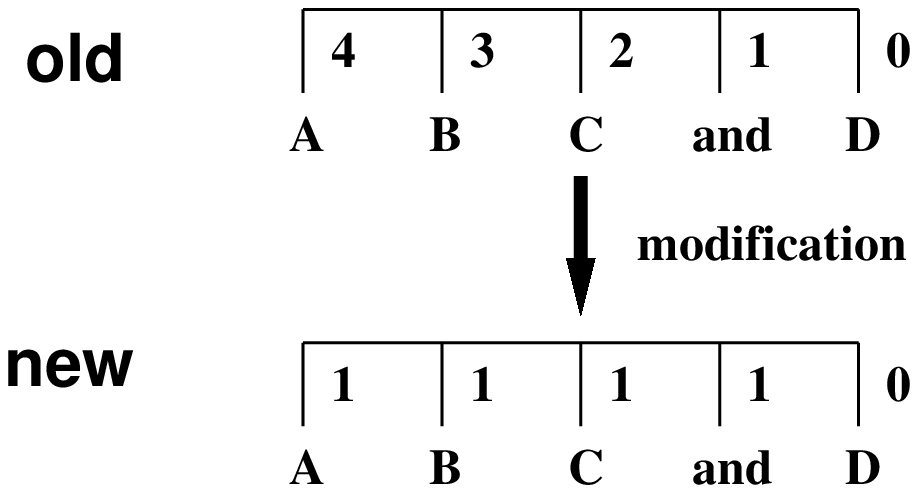} 
      \end{center}
      \vspace{-0.5cm}
    \caption{How to modify the number of each branch 
      in a coordination clause}
    \label{fig:eng_2}
    \end{minipage}
}
  \end{center}
\end{figure*}

\begin{table}[p]
  \caption{Number of nonterminals stored in a stack (Penn Treebank corpus)}
  \label{tab:hindo_toukei_eng_a}
  \begin{center}
\begin{tabular}[t]{|r|r|r|}
\multicolumn{3}{c}{Yngve's method in a word}\\\hline
\multicolumn{1}{|l|}{Stack} & \multicolumn{2}{c|}{Frequency}\\\cline{2-3}
   &   \multicolumn{1}{c|}{Words} & \multicolumn{1}{c|}{Sentences}\\\hline
0 & 49208 & 132\\
1 & 377740 & 772\\
2 & 309255 & 3921\\
3 & 213294 & 9528\\
4 & 103864 & 13324\\
5 & 44274 & 11163\\
6 & 16478 & 6158\\
7 & 5750 & 2719\\
8 & 1939 & 981\\
9 & 661 & 338\\
10 & 243 & 111\\
11 & 92 & 29\\
12 & 43 & 17\\
13 & 15 & 14\\
14 & 1 & 1\\\hline
\end{tabular}
\end{center}
\end{table}                                

\begin{table}[p]
  \caption{Number of nonterminals stored in a stack (Penn Treebank corpus)}
  \label{tab:hindo_toukei_eng_b}
  \begin{center}
\begin{tabular}[t]{|r|r|r|}
\multicolumn{3}{c}{Sampson's method in a word}\\\hline
\multicolumn{1}{|l|}{Stack}& \multicolumn{2}{c|}{Frequency}\\\cline{2-3}
     &   \multicolumn{1}{c|}{Words} & \multicolumn{1}{c|}{Sentences}\\\hline
0 & 49208 & 132\\
1 & 485849 & 1956\\
2 & 414945 & 13367\\
3 & 140611 & 22966\\
4 & 28317 & 9124\\
5 & 3616 & 1518\\
6 & 283 & 133\\
7 & 28 & 12\\\hline
\end{tabular}
\end{center}
\end{table}

\begin{table}[t]
  \caption{Number of nonterminals stored in a stack (Penn Treebank corpus)}
  \label{tab:hindo_toukei_eng_c}
  \begin{center}
\begin{tabular}[t]{|r|r|r|}
\multicolumn{3}{c}{Yngve's method in a NP}\\\hline
\multicolumn{1}{|l|}{Stack}& \multicolumn{2}{c|}{Frequency}\\\cline{2-3}
     &   \multicolumn{1}{c|}{NPs} & \multicolumn{1}{c|}{Sentences}\\\hline
0 & 69820 & 4546\\
1 & 102337 & 7634\\
2 & 74126 & 16847\\
3 & 30025 & 11489\\
4 & 11432 & 5780\\
5 & 3336 & 2020\\
6 & 963 & 633\\
7 & 273 & 187\\
8 & 76 & 51\\
9 & 29 & 13\\
10 & 13 & 8\\\hline
\end{tabular}

\end{center}
\end{table}

Using this method, 
Sampson analyzed 
the SUSANNE corpus (130,000 words) and 
obtained the results shown 
in Table \ref{tab:hindo_toukei_eng_sussane}(a) \cite{Sampson97}. 
``Frequency (words)'' means 
the frequency of words 
with the corresponding number of nonterminals stored in a stack. 
With this method of analysis 
many sentences exceeded the upper bound of $7 \pm 2$, i.e., 9. 
Sampson counted again, 
changing the number of each branch,  
as in Figure \ref{fig:eng_3}. 
With this new method, 
when A is recognized, 
B, C, D, and E are not remembered independently, 
but as one set of B, C, D, and E. 
Using this method, Sampson 
obtained the results shown in Table \ref{tab:hindo_toukei_eng_sussane}(b). 
This result showed that 
none of the sentences exceeded the lower bound of $7 \pm 2$, i.e., 5, 
therefore does not conflict with Miller's $7 \pm 2$ theory. 

We followed the same methods in an analysis of 
the corpus of The Wall Street Journal of 
Penn Treebank \cite{Marcus93}. 
We did not use the SUSANNE corpus because 
its structure is complicated, 
it is smaller than the Penn Treebank corpus, and 
it has already been studied by Sampson. 
The results for the Penn Treebank corpus 
are shown in Table \ref{tab:hindo_toukei_eng_a}. 
``Words'' means the frequency of words 
having a given number of nonterminals stored in the stack. 
``Sentences'' means 
the frequency of sentences 
having a given number of nonterminals stored in the stack. 
This time, we eliminated symbols such as periods, 
and we counted by changing the number of each branch 
in a coordination clause as in Figure \ref{fig:eng_2}, 
because the Penn Treebank corpus is constructed 
such that the extra number of nonterminals in a coordination clause 
is counted. 
The results in Table \ref{tab:hindo_toukei_eng_a} were 
found to match those in Table \ref{tab:hindo_toukei_eng_sussane}(a). 
Again, 
many sentences exceed the upper bound of seven plus or minus two. 
We also counted using Sampson's method. 
The results are shown in Table \ref{tab:hindo_toukei_eng_b}. 
Although the number of nonterminal symbols 
of the SUSANNE corpus did not exceed five, 
the Penn Treebank corpus 
included words with up to seven nonterminal symbols. 

We also developed a new counting method 
for an English corpus which is different from 
Yngve's and Sampson's methods. 
Our method is based on an idea that we should not use, 
as a cognitive unit, words but phrases, which corresponds to 
bunsetsus, which are the units for counting in Japanese. 
We assume that 
human beings recognize NPs all at once 
instead of dividing them into words, and 
count the number of nonterminals stored in a stack 
at the NP level. 
\None{
\footnote{If you compare the Japanese and English examples 
in Figures \ref{fig:jap} and \ref{fig:eng} carefully, 
you will notice that 
the Japanese phrases {\it sono} (the) and {\it shounen} (boy) or 
{\it chiisai} (small) and {\it ningyou} (doll) are divided 
and English phrases ``the boy'' and ``a small doll'' are 
not divided as noun phrases. 
You may think that 
a bunsetsu and a phrase are not the same cognitive units. 
But, we want you to see Figure \ref{fig:jap} again. 
In Figure \ref{fig:jap}, 
whether  {\it sono} (the) and {\it shounen} (boy) are grouped or divided, 
(and whether  {\it chiisai} (small) and 
{\it ningyou} (doll) are grouped or divided,) 
the maximum number of stored items is the same. 
So in Japanese we get the same result 
even using phrases as a unit, and the supposition 
that ``a phrase is a cognitive unit'' remains correct. 
Furthermore, we expect that even in English 
the recognition of a noun phrase does not use 
extra units of short term memory. 
In other words, an English noun phrase is 
expected to be recognized 
in the same manner as a Japanese phrase. 
This idea will be accepted by seeing that 
although English sentences use a right-branching structure, 
English noun phrases use a left-branching structure, 
the same as Japanese sentences.}. }%
In other words, 
we counted by using the sum of 
the numbers in the path from S to NP. 
The results shown in Table \ref{tab:hindo_toukei_eng_c}, 
are very similar to 
the results for Japanese sentences, 
shown in Table \ref{tab:hindo_toukei}, 
and contain sentences with eight and nine NPs, 
which correspond to the plus-two part of Miller's $7 \pm 2$ theory. 
These results show our method to be effective. 

Yngve's method did not obtain 
results that agree with Miller's $7 \pm 2$ theory, 
but Sampson's method and our method did. 
However, our method has the following two advantages 
over Sampson's method. 
\begin{itemize}
\item 
Our counting method in English, which uses bunsetsu-corresponding NPs  
as the unit for counting, 
is based on our counting method for Japanese. 
(It is plausible for several languages 
to have the same level of cognitive units.)

\item 
Although Sampson's method does not result in sentences 
with eight or nine nonterminal symbols, 
which is the upper bound of the $7 \pm 2$ theory, 
our method produced results that did. 
(Since ``$\pm 2$'' indicates an individual-based variation, 
a method that does not result in sentences with eight or nine 
nonterminals for a large corpus is very unnatural.) 

\end{itemize}

\section{Conclusion}

\None{
George A. Miller insisted that 
human beings have only 
seven chunks in short term memory plus or minus two. 
We counted the number of 
bunsetsus with undefined modifiees 
when investigating 
the dependencies from the beginning of Japanese sentences   
using the Kyoto University corpus, 
and we found that the number was roughly lower 
than nine, the upper bound of seven plus or minus two. 
We also investigated English sentences, 
and we got similar results to Japanese 
when we assumed that human beings 
recognize 
a series of words such as a noun phrase (NP) 
as a unit. 
This indicates that 
if we assume that the human cognitive units 
in Japanese and English are 
bunsetsu and NP respectively, 
analysis will support Miller's theory. }

We investigated Miller's $7 \pm 2$ theory 
using Japanese and English corpora. 
New information obtained in this paper is shown here. 
\begin{itemize}
\item 
  When bunsetsus were used as the cognitive unit, 
  the results of the investigation of 
  Japanese syntactic recognition 
  agreed with Miller's  $7 \pm 2$ theory. 
\item 
  When NPs were used as the cognitive unit, 
  the results of the investigation of 
  English syntactic recognition 
  agreed with Miller's  $7 \pm 2$ theory. 
  This indicates that 
  NPs are likely to be the cognitive unit. 
  It seems natural that 
  the NP level is the cognitive unit, 
  because it is the same level 
  as the Japanese cognitive unit, 
  bunsetsu\footnote{A cognitive unit is thought to be 
    a case element \cite{fillmore1} or a unit 
    taking the case element in the transformation process 
    from short-term memory to the semantic network of long-term memory. 
    So it seems natural that 
    the cognitive unit is the same level of phrase 
    in Japanese and English.}. 

\item 
  If we suppose that 
  bunsetsus and NPs are the cognitive units, 
  the analyses in Japanese and English support 
  Miller's  $7 \pm 2$ theory and also support
  Yngve's theory \cite{yngve60}, which is that 
  the number of 
  items stored in short-term memory does not exceed $7 \pm 2$ 
  in language understanding and generation. 
  These analyses support 
  Miller's  $7 \pm 2$ theory and Yngve's theory. 
  From the standpoint of natural language processing, 
  if Yngve's assertion is right, 
  the assertion that ``the number of items stored in short-term memory 
  does not exceed $7 \pm 2$'' can be used in 
  the construction of an practical NLP system. 
\end{itemize}

\bibliographystyle{plain}
\bibliography{mysubmit}

\end{document}